\documentclass[preprint,review,longtitle]{elsarticle}

\usepackage{amsmath,amssymb,amsfonts}
\usepackage{graphicx}
\usepackage{subcaption}
\usepackage{booktabs}
\usepackage{multirow}
\usepackage{textcomp}
\usepackage{hyperref}
\usepackage{array}
\usepackage{tabularx}
\usepackage{ragged2e}
\usepackage{threeparttable}
\usepackage{csquotes}
\usepackage{bbm}
\usepackage{algorithm}
\usepackage{algorithmic}
\usepackage{placeins}
\usepackage{enumitem}

\usepackage{caption}
\usepackage{etoolbox}
\usepackage{longtable}
\usepackage{multicol}
\usepackage{float}
\usepackage{graphicx}

\newcolumntype{L}[1]{>{\raggedright\arraybackslash}p{#1}}
\newcolumntype{Y}{>{\RaggedRight\arraybackslash}X}

\renewcommand{\arraystretch}{0.95}

\renewcommand{\thetable}{S\arabic{table}}
\renewcommand{\thefigure}{S\arabic{figure}}
\pretocmd{\longtable}{\small}{}{}

\begin{document}

\begin{frontmatter}

\title{DeepEN: A Deep Reinforcement Learning Framework for Personalized Enteral Nutrition in Critical Care}

\author[1]{Daniel J. Tan}
\author[2]{Jiayang Chen}
\author[3]{Dilruk Perera}
\author[2]{Kay Choong See\textsuperscript{†}}
\author[1]{Mengling Feng\textsuperscript{†,*}}

\address[1]{Institute of Data Science and Saw Swee Hock School of Public Health, National University of Singapore, Singapore}
\address[2]{National University Hospital, Singapore}
\address[3]{Saw Swee Hock School of Public Health, National University of Singapore, Singapore}

\begin{abstract}
\noindent\textbf{Objective:} 
Enteral nutrition (EN) delivery in the ICU remains suboptimal due to limited personalization and uncertainty regarding appropriate calorie, protein, and fluid targets under dynamic metabolic demands. We introduce \textit{DeepEN}, a reinforcement learning (RL) framework for personalized EN optimization using electronic health record data.

\noindent\textbf{Methods:} 
\textit{DeepEN} was trained on over 11{,}000 ICU patients from MIMIC-IV to generate 4-hourly, patient-specific caloric, protein, and fluid targets. The state representation incorporated demographics, comorbidities, vital signs, laboratory values, and recent interventions. A physiologically aligned reward framework balanced biomarker stability with long-term survival. Policy learning employed a dueling double deep Q-network with Conservative Q-Learning regularization to enable safe offline training.

\noindent\textbf{Results:}
\textit{DeepEN} achieved the highest estimated policy value (\(V^\pi = 9.48\)) and the lowest calibrated mortality (18.8 $\pm$ 1.0\%), representing a 4.0 percentage-point absolute reduction compared with clinician practice (22.8\%). The policy also demonstrated superior metabolic stability, achieving the highest proportion of glucose, phosphate, and sodium values within target range. Furthermore, deviation from the \textit{DeepEN} policy was independently associated with increased mortality and biomarker instability, whereas deviation from a random policy showed no such association. Interpretability analyses further indicated that recommendations were conditioned on physiologically relevant markers of organ function and metabolic status rather than static dosing heuristics.

\noindent\textbf{Conclusion:} 
\textit{DeepEN} demonstrates the feasibility of conservative offline RL for safe, individualized EN optimization, highlighting the potential of data-driven personalization to complement guideline-based approaches in critical care.
\end{abstract}
\end{frontmatter}

\vspace{-1em} 
\noindent\textsuperscript{†}\textit{Jointly supervised this work.}\\
\textsuperscript{*}\textit{Corresponding author: ephfm@nus.edu.sg}

\section{Introduction}

Enteral nutrition (EN) delivers nutrients directly to the gastrointestinal tract and is commonly initiated early during Intensive Care Unit (ICU) admission when oral intake is inadequate \cite{adeyinka2018enteric}. Critically ill patients frequently exhibit increased metabolic demand or impaired intake due to mechanical ventilation, neurological injury, gastrointestinal dysfunction, or hemodynamic instability \cite{preiser2021guide}. Although adequate caloric and protein provision is essential for recovery, determining appropriate dosing and timing remains clinically challenging.

Guidelines from the American Society for Parenteral and Enteral Nutrition (ASPEN) provide weight- and condition-specific intake targets \cite{compher2022guidelines}. However, randomized trials comparing aggressive versus conservative feeding strategies have reported heterogeneous findings \cite{compher2022guidelines,bels2024effect,heyland2023effect}, reflecting ongoing uncertainty regarding optimal nutritional intensity in critical illness. In practice, EN delivery often deviates from guideline targets due to evolving organ function, feeding tolerance, concurrent therapies, and rapidly changing metabolic demands \cite{mirhosiny2021physicians,friesecke2014improvement,jarden2015practice}. As a result, underfeeding, overfeeding, and delayed initiation remain common and are associated with adverse outcomes \cite{ramaswamy2024nine}.

These challenges highlight a central tension in critical care nutrition: while guidelines provide population-level structure, bedside decision-making is inherently dynamic, multifactorial, and patient-specific. Recent reviews in critical care nutrition have explicitly identified artificial intelligence (AI) and data-driven modeling as promising avenues for enabling more adaptive and individualized nutrition management in the ICU, particularly given the heterogeneity and rapidly evolving physiology of critically ill patients \cite{kittrell2024role,stoian2025personalized}. At the same time, these reviews emphasize that practical implementations remain limited, and methodological frameworks for translating high-dimensional ICU data into safe, interpretable, and clinically actionable nutrition policies are still underdeveloped.

The core challenge is therefore not merely personalization, but jointly optimizing interdependent nutritional components under delayed physiologic feedback and sparse long-term outcomes. This problem structure differs fundamentally from single-intervention titration tasks and requires methodological approaches that explicitly address multi-component action spaces, temporal credit assignment, and safety in observational data.

Reinforcement learning (RL) provides a natural framework for sequential clinical decision-making under uncertainty. By leveraging retrospective ICU trajectories, RL models can estimate policies that optimize long-term outcomes while conditioning decisions on evolving physiological states. While RL has demonstrated promise in domains such as sepsis management and mechanical ventilation optimization \cite{raghu2017deep,tan2024advancing,peine2021development,kaushik2022conservative}, its application to coordinated enteral nutrition management—characterized by multi-dimensional dosing decisions and delayed metabolic effects—remains unexplored. To our knowledge, no prior study has developed and systematically evaluated a reinforcement learning framework for dynamic enteral nutrition personalization in critical care.

In this study, we introduce DeepEN, a conservative offline reinforcement learning framework for dynamic, patient-specific optimization of enteral nutrition therapy in the ICU.

Our main contributions are as follows:

\begin{itemize}
    \item We formulate enteral nutrition management as a multi-component, combinatorial sequential decision problem with delayed physiologic feedback, and develop a conservative offline reinforcement learning framework tailored to this class of delayed, multi-component treatment problems.
    
    \item We design a physiologically aligned reward framework for slow-acting metabolic therapies, combining terminal survival with intermediate biomarker-stability signals modeled using a reverse-Huber (BerHu) formulation evaluated over forward windows aligned with metabolic response times.

    \item We introduce a structured evaluation framework that links learned value estimates to empirical mortality and biomarker stability through return–outcome calibration, complemented by policy-level structural analyses and interpretability assessments to characterize decision coherence and safety.
\end{itemize}

\noindent
\begin{minipage}{\textwidth}

\textbf{\large Statement of Significance}

\vspace{0.5em}

\noindent
\begin{tabular}{p{0.28\textwidth} p{0.68\textwidth}}
\toprule
\textbf{Problem or Issue} &
Enteral nutrition (EN) dosing in the ICU is highly variable and difficult to individualize due to interdependent calorie, protein, and fluid decisions with delayed metabolic effects and sparse outcome feedback. \\[0.6em]

\textbf{What is Already Known} &
Guideline-based EN strategies rely primarily on static, weight-based heuristics and do not adapt to dynamic physiology. Although reinforcement learning has been applied to other ICU domains (e.g., sepsis management), coordinated multi-component nutritional optimization with delayed feedback has not been systematically addressed. \\[0.6em]

\textbf{What This Paper Adds} &
We introduce \textit{DeepEN}, a conservative offline reinforcement learning framework for dynamic, patient-specific EN optimization. The study establishes a reproducible methodological approach for multi-component, slow-acting clinical therapies, integrating physiologically aligned reward design, structured return–outcome calibration, and policy-level interpretability analyses to promote safety and clinical plausibility. \\[0.6em]

\textbf{Who Would Benefit from the Knowledge in This Paper} &
Clinical informatics researchers, critical care clinicians, and health system developers seeking rigorous, safety-conscious approaches to AI-driven decision support for complex, longitudinal ICU therapies. \\
\bottomrule
\end{tabular}

\end{minipage}
\vspace{1.0em}

\section{Background and Related Work}

\subsection{\textbf{Reinforcement Learning}}

Reinforcement learning (RL) provides a framework for optimizing sequential decision-making under uncertainty and is commonly formalized as a Markov Decision Process (MDP), defined by the tuple \( (\mathcal{S}, \mathcal{A}, r, P, \gamma) \), where \( \mathcal{S} \) denotes the state space, \( \mathcal{A} \) the action space, \( r \) the reward function, \( P \) the transition dynamics, and \( \gamma \in (0,1) \) the discount factor. At each time step \( t \), an agent observes state \( s_t \), selects action \( a_t \), transitions to \( s_{t+1} \sim P(\cdot \mid s_t, a_t) \), and receives reward \( r_t \).

The objective of RL is to learn a policy \( \pi \) that maximizes the expected cumulative discounted return \( \mathbb{E}[\sum_{t=0}^{T} \gamma^t r_t] \). In healthcare, this framework naturally aligns with dynamic treatment regimes, where actions correspond to treatment decisions and rewards encode short- and long-term clinical outcomes.

\subsection{\textbf{Q-Learning and Offline Reinforcement Learning}}

Q-learning \cite{watkins1992q} is a foundational RL algorithm that estimates the action-value function \( Q(s,a) \), representing the expected return of taking action \( a \) in state \( s \). The Q-function is updated via the Bellman equation:

\begin{equation}
Q(s_t, a_t) \leftarrow Q(s_t, a_t) + \eta \left( r_t + \gamma \max_{a'} Q(s_{t+1}, a') - Q(s_t, a_t) \right),
\end{equation}

where \( \eta \) is the learning rate. 

For high-dimensional clinical state spaces, Deep Q-Networks (DQN) \cite{mnih2015human} approximate \( Q(s,a) \) using neural networks. However, the maximization operator in the Bellman target can induce overestimation bias, particularly when Q-values are noisy or poorly supported by data.

This issue is amplified in \emph{offline} (batch) reinforcement learning \cite{levine2020offline}, where policies are learned from a fixed retrospective dataset without further environment interaction. In healthcare, offline RL is essential, as online exploration on real patients is infeasible and unethical. However, learning from static data introduces distributional shift: state–action pairs not well represented in the dataset may yield unreliable Q-value estimates. In high-stakes settings, such errors can translate into unsafe treatment recommendations.

\subsection{\textbf{Mitigating Overestimation in Offline RL: D3QN and CQL}}

To address overestimation bias, Double Q-learning decouples action selection from evaluation, reducing upward bias in Q-targets \cite{van2016deep}. The Dueling Double DQN (D3QN) architecture further improves stability by separating state-value and advantage estimation \cite{wang2016dueling,hessel2018rainbow}. This structure enhances learning efficiency and robustness, particularly in complex clinical state spaces.

While D3QN mitigates maximization bias, it does not explicitly address distributional shift in offline settings. Conservative Q-Learning (CQL) \cite{kumar2020conservative} introduces an additional regularization term to penalize overestimated Q-values for actions that are not well supported by the dataset. The CQL objective augments the standard Bellman error with a conservative penalty:

\begin{equation}
\mathcal{L}_{\text{CQL}} =
\underbrace{\mathcal{L}_{\text{Bellman}}}_{\text{TD error}}
+ \alpha_{\text{CQL}} \left(
\mathbb{E}_{s \sim \mathcal{D}, a \in \mathcal{A}}[Q(s,a)]
-
\mathbb{E}_{(s,a) \sim \mathcal{D}}[Q(s,a)]
\right),
\end{equation}

where the first expectation increases penalties for high Q-values assigned to arbitrary actions, and the second maintains higher values for actions observed in the dataset. This encourages pessimistic value estimates for out-of-distribution actions while preserving support for clinician-like decisions.

CQL is particularly important in our application due to the combinatorial structure of the action space. Each decision jointly specifies caloric, protein, and water dosing levels, resulting in a substantially larger discrete action space than typical single-intervention problems. Many state–action combinations are therefore sparsely represented in retrospective data. Without conservative regularization, the model may overestimate the value of rarely observed nutritional combinations, potentially generating unsafe or clinically implausible recommendations. By penalizing unsupported actions, CQL helps constrain the learned policy to remain within regions of the state–action space supported by observed ICU practice, and it has successfully been applied to other treatment optimization applications in the past \cite{kaushik2022conservative}. In this work, we combine D3QN with CQL regularization to promote stable value learning while maintaining safety and clinical plausibility in a fully offline setting.

\subsection{\textbf{Reinforcement Learning for Enteral Nutrition Management}}

Reinforcement learning has been extensively studied for critical care applications, including fluid and vasopressor dosing in sepsis \cite{raghu2017deep,tan2024advancing}, mechanical ventilation management \cite{peine2021development}, and sedation titration \cite{yu2021reinforcement}. However, its application to enteral nutrition (EN) management remains unexplored.

Existing AI efforts in nutrition have primarily focused on malnutrition screening \cite{besculides2023implementing,raphaeli2023using} or predicting feeding intolerance \cite{hu2022development,chen2023development}, rather than sequential optimization of dosing strategies. 

EN presents unique methodological challenges. Unlike hemodynamic interventions that produce rapid physiological effects, nutritional interventions exert gradual and diffuse metabolic impacts, often confounded by concurrent therapies and heterogeneous patient trajectories. These characteristics complicate reward design, credit assignment, and outcome attribution. 

Accordingly, effective RL-based EN management requires carefully constructed state representations and reward functions grounded in clinical expertise and physiologic insight. Our work represents the first application of conservative offline RL to personalized enteral nutrition optimization in critical care, addressing both methodological and domain-specific challenges inherent to this setting.

\section{Methods}
\begin{figure}[!hbt]
\centering
\includegraphics[width=1\columnwidth]{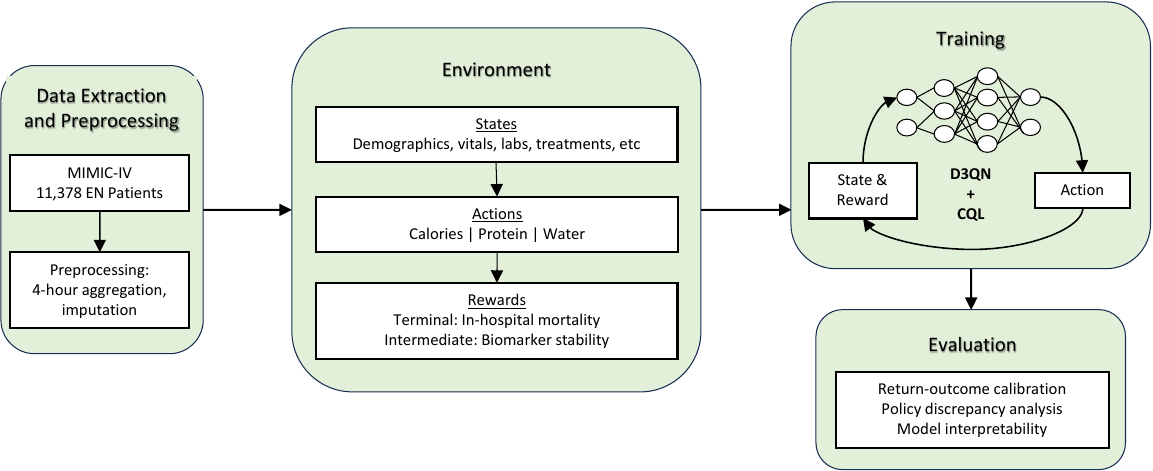} 
\caption{Overview of the DeepEN solution pipeline.}
\label{fig:overview}
\vspace{-0.5em} 
\end{figure}

This section details our methods, including RL problem formulation, algorithm design, cohort construction and experiment setup. Figure \ref{fig:overview} illustrates an overview of the solution pipeline.

\subsection{RL Problem Definition}
\subsubsection{States}
Our state space (S) comprises 102 variables consisting of 63 base variables , and an additional 39 ‘rate-of-change’ variables. (Table \ref{tab:included_variables}) The rate-of-change variables are the average rate of change calculated across the previous three time-steps  for 39 chosen base variables. These variables capture clinically meaningful temporal trends and provide additional contextual information regarding disease progression and physiologic trajectory.

{\renewcommand{\arraystretch}{0.6} 

\begin{table}[!t]
\caption{State Space Variables}
\label{tab:included_variables}
\centering
\footnotesize
\begin{threeparttable}
\setlength{\tabcolsep}{3pt} 
\begin{tabular}{p{2.8cm}p{6.5cm}}
\toprule
\textbf{Category} & \textbf{Variables} \\
\midrule
Demographics (4) & Age, Gender, Weight\textsuperscript{*}, ICU readmission \\
Diagnoses and Comorbidities (6) & Burns, CKD, Diabetes, Sepsis, Trauma, Elixhauser score \\
Vitals (13)\textsuperscript{*} & HR, SBP, MBP, DBP, Resp. rate, Temperature, PaCO\textsubscript{2}, PaO\textsubscript{2}, PF ratio, SpO\textsubscript{2}, SOFA, GCS, Shock index \\
Labs (24)\textsuperscript{*} & Albumin, pH, Calcium, Glucose, Hemoglobin, Magnesium, WBC, Creatinine, Bicarbonate, Sodium, Lactate, Chloride, Platelets, Potassium, PTT, PT, AST, ALT, BUN, INR, Ionised calcium, Total bilirubin, Base excess, Phosphate \\
Feeding Related (5) & Prev. calories, Prev. protein, Prev. water, Cumulative calories, Cumulative protein \\
Treatments and Interventions (8) & Mechanical ventilation, FiO\textsubscript{2}, CRRT, Prev. IV fluids, Vasopressor dose, Propofol dose, Insulin dose, 24-h cumulative insulin dose \\
Others (3) & Urine output (4-hourly)\textsuperscript{*}, Total output, Time since EN initiation \\
\bottomrule
\end{tabular}

\vspace{0.5em}
\raggedright
\textsuperscript{*}\,Variables whose average rate of change was also included.\\[2pt]
\textbf{Abbreviations:} CKD = chronic kidney disease; DBP = diastolic blood pressure;  
GCS = Glasgow Coma Scale; HR = heart rate; ICU = intensive care unit;  
MBP = mean blood pressure; PTT = partial thromboplastin time;  
Resp. = respiratory rate; SBP = systolic blood pressure; WBC = white blood cell.
\end{threeparttable}
\end{table}

\subsubsection{Actions}
The action space is defined using three main components of enteral nutrition: calories, protein and water. 4-hourly decision intervals were chosen to align with ICU nutrition adjustment practices and the sampling frequency of key physiological variables, while providing sufficient temporal resolution and trajectory length for effective offline reinforcement learning. For each patient and time window, we consider the weight-adjusted quantities of these components, each discretized into four levels based on empirical quantiles. In line with clinical guidelines, we define total energy and fluid intake as the sum of EN and non-EN sources (e.g., IV fluids and propofol). Although non-EN sources are included for completeness, EN make up the vast majority of these nutritional components.

The action space A is the Cartesian product of the set of these three components, resulting in a theoretical total of 64 ($4\times4\times4$) possible action combinations. However, due to inherent correlation among calories, protein and water administration, only 51 of the combinations are observed in the dataset. Hence these combinations are considered as the  functional action space. Note that the quantities of all three components are strictly positive, as we focus only on periods where enteral nutrition was actively administered. Further details on the action composition and discretization process is detailed in Supplementary Table S1 and S3.

\subsubsection{Rewards}

We define a reward function consisting of intermediate biomarker rewards and a terminal mortality reward, following established reinforcement learning formulations in critical care while tailoring the feedback signals to the physiological dynamics of enteral nutrition.

Let $R_t \in \mathbb{R}$ denote the scalar reward at time step $t$. For each trajectory:
\begin{equation}
R_t =
\begin{cases}
R_{\text{term}}(m), & d_t = 1 \\
R_{\text{bio},t}, & d_t = 0
\end{cases}
\label{eq:reward_definition}
\end{equation}
where $d_t \in \{0,1\}$ indicates terminal status and $m \in \{0,1\}$ denotes ICU mortality.

The terminal reward reflects survival outcome:
\begin{equation}
R_{\text{term}}(m) =
\begin{cases}
+r_T, & \text{if } m = 0 \\
-r_T, & \text{if } m = 1
\end{cases}
\label{eq:terminal_reward}
\end{equation}

We set $r_T = 15$, consistent with prior clinical RL studies \cite{raghu2017deep,tan2024advancing}, ensuring survival remains the dominant optimization objective.

\vspace{0.5em}

Intermediate nutrition-sensitive biomarker rewards ($R_{\text{bio}}$) mitigate reward sparsity and improve temporal credit assignment by providing physiologically interpretable feedback throughout the ICU stay. Because laboratory markers respond to nutrition over clinically distinct physiological timescales, a single next-state formulation would introduce temporal mismatch between actions and observable effects. We therefore compute rewards using biomarker-specific forward evaluation windows, defined as measurements occurring after the current decision time. 

Within each window, the biomarker value is represented by the mean measurement across the window. This aggregation approximates the sustained physiological response to nutritional exposure rather than a single laboratory snapshot, improving temporal alignment between intervention and metabolic effect while reducing sensitivity to measurement noise, irregular sampling frequency, and missing laboratory draws.

We define the selected biomarkers, their clinically accepted maintenance ranges, and their physiologically motivated evaluation windows in Table~\ref{tab:biomarker_windows}. These markers collectively capture short-term carbohydrate tolerance, electrolyte shifts during nutritional repletion, and fluid balance stability.

\begin{table}[ht]
\centering
\footnotesize
\setlength{\tabcolsep}{5pt}
\renewcommand{\arraystretch}{1.25}
\caption{Biomarker Target Ranges and Physiologically Motivated Evaluation Windows}
\begin{tabular}{l c c p{5.2cm}}
\toprule
\textbf{Biomarker} & \textbf{Target Range} & \textbf{Evaluation Window} & \textbf{Physiologic Relevance} \\
\midrule

\textbf{Glucose}    
& $[100,\,180]$ mg/dL$^{\cite{wu2022expert}}$ 
& 4 h  
& Rapid response to carbohydrate delivery and insulin modulation; reflects short-term metabolic tolerance to feeding rate and glycemic safety margins$^{\cite{gosmanov2013management,rice2019dietary}}$  \\[6pt]

\textbf{Phosphate}  
& $[2.5,\,4.5]$ mg/dL$^{\cite{geerse2010treatment}}$ 
& 8--16 h  
& Insulin-driven intracellular phosphate uptake during nutritional repletion; early indicator of refeeding syndrome and anabolic activation$^{\cite{kalantar2010understanding,mehanna2008refeeding,da2020aspen}}$  \\[6pt]

\textbf{Sodium}     
& $[135,\,145]$ mmol/L$^{\cite{braun2015diagnosis}}$ 
& 12--24 h  
& Reflects water balance and osmotic homeostasis; captures hydration adequacy and consequences of fluid dosing decisions$^{\cite{van2016deep, besen2015fluid}}$ \\

\bottomrule
\end{tabular}
\label{tab:biomarker_windows}
\end{table}

Rather than rewarding only threshold violations, we directly penalize the magnitude of physiological deviation from the desired metabolic state. This formulation reflects clinical reasoning: small deviations are often tolerated, but large abnormalities represent progressively greater physiological risk. To encode this property, we adopt a reverse-Huber (BerHu) loss formulation, which applies linear penalties near the target range and quadratic penalties for large deviations.

For biomarker $x$ with target range $[x_{\min}, x_{\max}]$, define the normalized deviation from the midpoint as
\begin{equation}
u = \frac{x - c}{w},
\quad
c = \frac{x_{\min} + x_{\max}}{2}, \quad
w = \frac{x_{\max} - x_{\min}}{2}.
\label{eq:normalized_dev}
\end{equation}

The BerHu loss is defined as
\begin{equation}
\mathcal{L}_{\text{BerHu}}(u) =
\begin{cases}
|u|, & |u| \le 1 \\
\frac{u^2 + 1}{2}, & |u| > 1
\end{cases}
\label{eq:berhu}
\end{equation}

The biomarker reward is then
\begin{equation}
R_x = 1 - \mathcal{L}_{\text{BerHu}}(u).
\label{eq:berhu_reward}
\end{equation}

This produces a reward landscape where:
(i) values near the center of the target range receive maximal reward,
(ii) mild deviations are penalized gently,
and (iii) severe abnormalities incur increasingly strong penalties.
Clinically, this mirrors metabolic management goals — maintaining stability while strongly discouraging dangerous derangements such as severe dysglycemia, hypophosphatemia during refeeding, or dysnatremia from fluid imbalance.

The composite biomarker reward is
\begin{equation}
R_{\text{bio}} = \sum_{x \in \mathcal{B}} \rho_x R_x,
\label{eq:bio_composite}
\end{equation}
where $\mathcal{B}$ denotes the set of biomarkers and weights $\rho_x$ control the relative contribution of each physiologic signal. In this study, all weights were set equal to avoid imposing subjective prioritization among physiologic systems in the absence of strong clinical evidence favoring one biomarker over another. More generally, the weighting formulation permits domain-specific adjustment in other treatment optimization settings where certain physiologic targets may warrant greater emphasis based on clinical risk, therapeutic priorities, or expert knowledge.

Because biomarkers operate on different numerical scales and clinical variances, each biomarker reward is normalized by its median absolute magnitude across non-terminal steps prior to aggregation. This ensures each physiologic system contributes comparably to the learning signal rather than allowing inherently higher-variance markers to dominate optimization.

To ensure that intermediate shaping rewards do not overshadow the terminal survival objective, we apply trajectory-level scaling.

For each trajectory $i$, the cumulative intermediate reward is defined as:
\begin{equation}
S_i = \sum_{t:d_t = 0} R^{\text{bio}}_{i,t}
\label{eq:trajectory_sum}
\end{equation}

We define a scaling factor:
\begin{equation}
s = \frac{\lambda r_T}{\mathrm{median}_i(|S_i|)}
\label{eq:scaling_factor}
\end{equation}
where $\lambda$ controls the relative contribution of intermediate rewards to the terminal outcome. 

Intermediate rewards are multiplied by $s$ prior to terminal assignment. This constrains the median cumulative shaping magnitude to approximately $\lambda$ times the terminal reward, preserving mortality as the dominant optimization objective while retaining clinically meaningful physiologic guidance, consistent with prior reinforcement learning dynamic treatment regime studies \cite{raghu2017deep,wu2023value}.

To assess robustness to this design choice, we conducted a sensitivity analysis over 
$\lambda \in \{0, 0.1, 0.2, 0.3, 0.4, 0.5\}$, including $\lambda = 0$ (i.e., no intermediate rewards). Policy performance across these settings is reported in Supplementary Table~S5. We found that $\lambda = 0.2$ achieved the greatest mortality reduction and performance robustness under resampling. Smaller values reduced the influence of physiologic shaping, whereas larger values increasingly shifted optimization away from the terminal survival objective.

\subsection{Experimental Setup}

\subsubsection{Data Extraction and Preprocessing}
We extracted a cohort of enteral feeding patients from the MIMIC-IV database \cite{johnson2023mimic} which contains data for over 60,000 ICU patients from the Beth Israel Deaconess Medical Center (Boston, MA, USA) from between 2008 to 2019. Inclusion criteria included the following conditions: 1) patient is $>$18 years of age, 2) has at least 12 hours of enteral feeding data, and 3) has at least one recorded weight measurement. In total, data for 11,378 patients were included. For each patient, we collect data on demographics, vital signs, lab-values, treatments, specific diagnoses, enteral feeding data, and other relevant variables detailed in Table \ref{tab:included_variables}. We only include data from periods where EN is administered, and did not consider periods where EN is absent or interrupted, nor did we consider data from parenteral feeding. In line with clinical evidence, we use enteral feeding data only from the first 10 days post-ICU admission to focus on the acute phase of critical illness (PCI) \cite{iwashyna2015towards}. Data for calories, protein, and water intake were extracted from MIMIC-IV's \textit{inputevents} table. We aggregated the patient trajectories into four-hourly windows using mean or sum as appropriate. Missing values were imputed using sample-and-hold with time-since-last-measurement features. More details on the patient cohort can be found in Supplementary Table~S2.

\subsubsection{Baselines}
To benchmark the performance of our reinforcement learning (RL) policy, we compare it against four distinct baselines: a random dosage policy, a clinician policy, a behavior cloning (BC) policy, and an ASPEN-derived policy.

\begin{enumerate}
    \item \textbf{Random Dosage Policy:} The random policy selects actions uniformly at random from the discrete action space. Formally, it is defined as $\pi(a) = \frac{1}{M}$, where $M$ is the total number of possible actions. This baseline serves as a critical sanity check, establishing a worst-case scenario or lower performance bound. It allows us to assess whether the learned policy outperforms naive, uninformative behavior.
    
    \item \textbf{Clinician Policy:} The clinician policy directly reflects the observed actions taken by human clinicians in the MIMIC-IV dataset. It represents real-world practice and serves as an empirical reference point for evaluating the clinical plausibility of the learned policy.
    
    \item \textbf{Behavior Cloning (BC) Policy:} The BC policy is a supervised learning model trained to imitate clinician decisions. Specifically, it is implemented as a neural network that minimizes the cross-entropy loss between the predicted actions and those recorded in the dataset. Although it may not replicate clinician behavior perfectly, its performance is expected to be similar to the clinician policy. This baseline helps evaluate whether the use of reinforcement learning yields meaningful improvements over simpler supervised learning approaches.
    
    \item \textbf{ASPEN-derived Policy:} This rule-based policy encodes clinical recommendations drawn from the American Society for Parenteral and Enteral Nutrition (ASPEN) guidelines \cite{compher2022guidelines}, in combination with more recent evidence-based literature \cite{wischmeyer2023personalized,stoian2025personalized}, to reflect up-to-date, evidence-informed recommendations for enteral nutrition in critically ill patients. The policy is deterministic and constructed according to expert-derived nutritional targets and clinical rules. A full specification of this policy is provided in the Supplementary Information (Supplementary Methods: \textit{ASPEN-Derived Policy Definition}).
\end{enumerate}

\subsubsection{Training and Hyperparameters}

Patient trajectories were randomly partitioned at the patient level into 70\% training, 10\% validation, and 20\% testing sets. Hyperparameter tuning and model selection were performed on the validation set using CWPDIS-based offline evaluation, while all final performance metrics reported in the manuscript were evaluated exclusively on the held-out test set.

Our deep RL model is based on the D3QN framework with CQL regularization. A grid search was conducted over the CQL coefficient $\alpha \in \{0.01, 0.1, 0.5, 1\}$ and discount factor $\gamma \in \{0.75, 0.9, 0.95, 0.99\}$. We additionally evaluated ReLU and sigmoid activations with 1--4 hidden layers and layer widths selected from $\{64, 128, 256, 512\}$. All models were trained using a learning rate of $1\times10^{-4}$ and batch size of 500.

The final model achieved optimal validation-set performance with $\alpha = 0.5$, $\gamma = 0.99$, ReLU activation, and a 3-layer architecture with dimensions 256, 128, and 64.

\subsubsection{Off-Policy Evaluation}

In online reinforcement learning (RL), policies are learned and evaluated via direct interaction with the environment. However, in the healthcare RL context—where the "environment" corresponds to real patients—any real-time exploration is infeasible. Instead, we evaluate the performance of policies using various quantitative and qualitative off-policy evaluation (OPE) metrics tailored specifically to our target clinical outcomes. These do not require a simulator and instead adopt a well-established approach using offline learning on retrospective healthcare data.

\begin{enumerate}

\item \textbf{Return-Calibrated Off-Policy Evaluation:}

We use a doubly robust (DR) off-policy evaluation framework \cite{jiang2016doubly} to estimate the expected discounted return \(V^\pi\) of each evaluated policy. DR estimation combines importance sampling with learned value-function approximation to reduce the variance and instability commonly observed in pure importance-sampling approaches, and has been widely adopted in prior healthcare RL studies \cite{wu2022expert,lu2024reinforcement,gottesman2018evaluating,sun2021personalized,weng2017representation}. To estimate policy-specific value functions, we trained fitted Q evaluation (FQE) models conditioned on each evaluated policy \cite{le2019batch}. Trajectory-level DR estimates were then computed and averaged across observed clinician trajectories to obtain the final policy value \(V^\pi\).

To obtain clinically interpretable outcome estimates, we constructed empirical calibration relationships between learned value estimates and observed outcomes under the clinician policy. Specifically, trajectories were grouped into discrete return bins according to trajectory-level mean SARSA-estimated \(Q\)-values derived from observed clinician actions, and the corresponding observed in-hospital mortality and biomarker percent-in-range (PIR) were computed within each bin. Because these trajectories were generated under the clinician policy, the associated mortality and biomarker outcomes correspond to directly observed ground-truth outcomes in the retrospective dataset. A well-learned value function should induce a monotonic relationship whereby higher predicted returns correspond to lower mortality and improved physiologic stability.

After establishing these empirical return–outcome calibration curves, \(V^\pi\) for DeepEN, ASPEN, BC, and Random policies were mapped onto the calibrated return–mortality and return–PIR functions to obtain clinically interpretable outcome estimates. Returns were therefore used as an ordering and calibration signal rather than as direct probability predictions, thereby avoiding explicit counterfactual outcome simulation while grounding interpretation in empirically observed clinician trajectories.

\vspace{0.3em}

\item \textbf{Policy Discrepancy and Outcome Structure:}

To characterize how disagreement between policies and clinician decisions relates to outcomes, we examined the empirical relationship between action-level differences (clinician $-$ policy) and clinical endpoints using the U-curve framework \cite{raghu2017deep,gottesman2018evaluating}.

Discrete enteral nutrition actions were decoded into calorie, protein, and water components. At each decision step, action-level discrepancy was defined as the absolute difference between clinician and evaluated policy recommendations. Discrepancies were binned, and mortality rates or forward-window biomarker deviations were computed within each bin to visualize how outcomes vary as policies diverge from observed clinician behavior.

While these discrepancy–outcome curves provide intuitive structural insights, one may argue that trajectories with low discrepancy (near zero) correspond to inherently healthier patients. To address this potential confounding, we performed multivariable regression analyses adjusting for baseline severity.

For ICU stay $i$, per-step discrepancy was defined as:

\begin{equation}
d^{(i)}_t = \frac{1}{3} \sum_{k \in \{\mathrm{cal},\mathrm{pro},\mathrm{wat}\}} 
\left| \hat{A}^{(i)}_{k,t} - A^{(i)}_{k,t} \right|,
\end{equation}

and averaged across time to obtain a stay-level mean policy distance. Logistic regression (mortality) and linear regression (biomarker deviation) models were fit with mean policy distance as the primary covariate, adjusting for age, gender, SOFA score, and weight. This analysis evaluates whether deviation from a given policy is independently associated with adverse outcomes after accounting for baseline clinical risk.

\vspace{0.3em}

\item \textbf{Model Interpretability:}

To assess the clinical interpretability of clinician and RL-derived nutrition policies, we conducted two complementary analyses:

\begin{enumerate}[label=(\roman*)]

\item \textit{Feature importance analysis:}

For each nutrition domain (calories, protein, water), separate random forest classifiers were trained to predict the discrete action selected at each decision interval under either the clinician policy or the DeepEN policy. Feature importance was quantified using permutation importance. Autoregressive and trivially predictive variables (e.g., prior nutrition delivery levels and cumulative nutrition variables) were excluded to focus on clinically meaningful drivers of decision-making.

\item \textit{Joint action distribution analysis:}

We constructed pairwise heatmaps of joint action distributions across calorie, protein, and water dosing levels to compare clinician and DeepEN treatment strategies. Heatmaps were stratified into early enteral nutrition (EN) phase ($\leq$3 days) and late EN phase ($>$3 days), reflecting common clinical practice patterns of early permissive underfeeding followed by progressive escalation.

\end{enumerate}

\end{enumerate}

\section{Results}

We evaluated DeepEN against four comparator policies: observed clinician practice, a random policy, behavior cloning (BC), and a guideline-based ASPEN policy. Evaluation was conducted using complementary quantitative and structural analyses, including off-policy value estimation, return-calibrated mortality and biomarker outcomes, and discrepancy-based assessments of policy behavior.

\subsection{Quantitative Off-Policy Evaluation}

Figure~\ref{fig:qplot} shows a consistent inverse relationship between expected return and observed in-hospital mortality under the clinician policy. Mortality declines steadily with increasing return, with the steepest gradient observed in the higher-value regions. Adequate support across return bins indicates that this trend is stable rather than driven by sparsely sampled trajectories. Together with the return–biomarker calibration curves (Supplementary Figure~S1), which demonstrate increasing percent-in-range with higher return across glucose, phosphate, and sodium, these findings support that the learned value function meaningfully stratifies trajectories by both survival risk and physiologic stability. Notably, the calibrated clinician mortality estimate (22.8\%) closely approximated the empirically observed mortality rate in the held-out test set (22.5\%), providing an additional sanity check on the fidelity of the calibration framework.

\begin{figure}[!hbt]
\centering
\includegraphics[width=0.75\columnwidth]{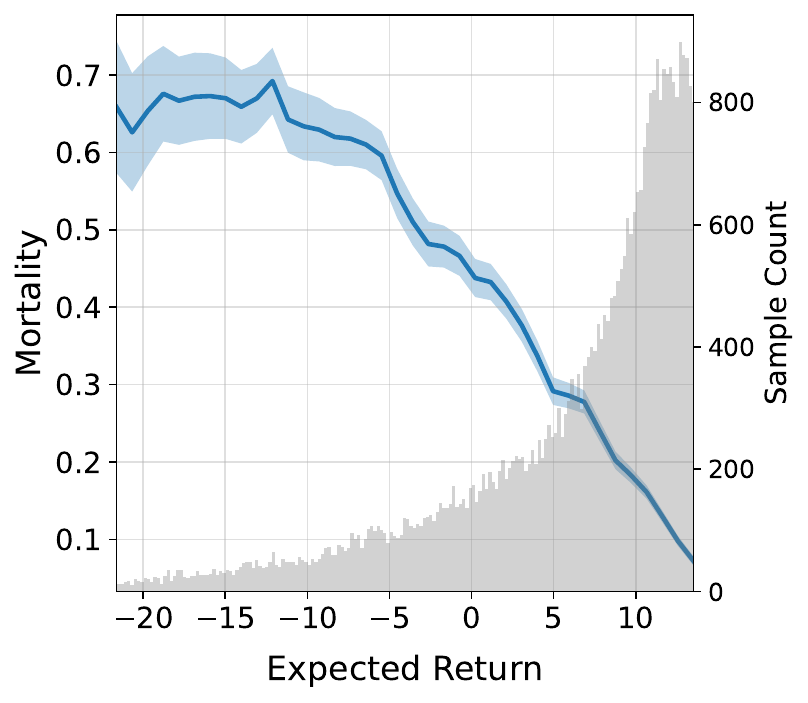} 
\caption{Observed in-hospital mortality as a function of expected returns, computed from clinician-policy trajectories in the held-out test set. The solid curve shows the estimated mortality rate across return bins, with shaded regions indicating 95\% confidence intervals. The overlaid histogram depicts the empirical distribution of expected returns across patient trajectories.}
\label{fig:qplot}
\vspace{-0.5em} 
\end{figure}

\begin{table*}[ht]
\centering
\normalsize
\caption{Comparative Policy Evaluation: Expected Return, Mortality, and Biomarker Stability (values reported as mean $\pm$ 95\% CI half-width)}
\resizebox{\textwidth}{!}{
\begin{tabular}{lccccc}
\toprule
\textbf{Policy} 
& \textbf{$V^\pi$} 
& \textbf{Mortality (\%)} 
& \textbf{Glucose \%} 
& \textbf{Phosphate \%} 
& \textbf{Sodium \%} \\
\midrule

Clinician 
& 8.11 $\pm$ 0.18
& 22.8 $\pm$ 1.3
& 72.1 $\pm$ 1.4
& 65.3 $\pm$ 1.5
& 62.6 $\pm$ 1.5 \\

Random    
& 5.73 $\pm$ 0.93
& 28.7 $\pm$ 1.7
& 65.8 $\pm$ 1.8
& 64.8 $\pm$ 1.8
& 57.9 $\pm$ 1.9 \\

BC      
& 8.17 $\pm$ 0.19
& 22.6 $\pm$ 1.3 
& 72.3 $\pm$ 1.4
& 65.4 $\pm$ 1.5 
& 62.7 $\pm$ 1.5 \\

ASPEN    
& 7.52 $\pm$ 0.65
& 25.2 $\pm$ 1.4
& 70.4 $\pm$ 1.5 
& 64.3 $\pm$ 1.6
& 61.4 $\pm$ 1.6 \\

DeepEN    
& \textbf{9.48 $\pm$ 0.30} 
& \textbf{18.8 $\pm$ 1.0} 
& \textbf{76.9 $\pm$ 1.1} 
& \textbf{67.7 $\pm$ 1.3} 
& \textbf{65.2 $\pm$ 1.3} \\

\bottomrule
\end{tabular}
}

\label{tab:quant_metrics}
\end{table*}

Table~\ref{tab:quant_metrics} summarizes calibrated policy outcomes across clinician, Random, BC, ASPEN, and DeepEN policies. DeepEN achieved the highest policy value \(V^\pi\) and the lowest estimated mortality (18.8\%), outperforming both clinician practice (22.8\%) and guideline-based ASPEN (25.2\%). This corresponds to an absolute mortality reduction of 4.0\% relative to clinician care—approximately 40 fewer deaths per 1000 ICU patients. DeepEN also demonstrated the strongest metabolic stability, achieving the highest proportion of glucose, phosphate, and sodium measurements within target range.

The Random policy performed worst across all metrics, whereas behavior cloning closely mirrored clinician performance, consistent with its imitation objective. Clinician practice outperformed ASPEN across mortality and biomarker endpoints, likely reflecting clinicians’ adaptation to dynamic physiologic context beyond static rule-based targets.

\subsection{Policy Discrepancy and Outcome Structure}

\begin{figure*}[!t]
\centering
\includegraphics[width=1\textwidth]{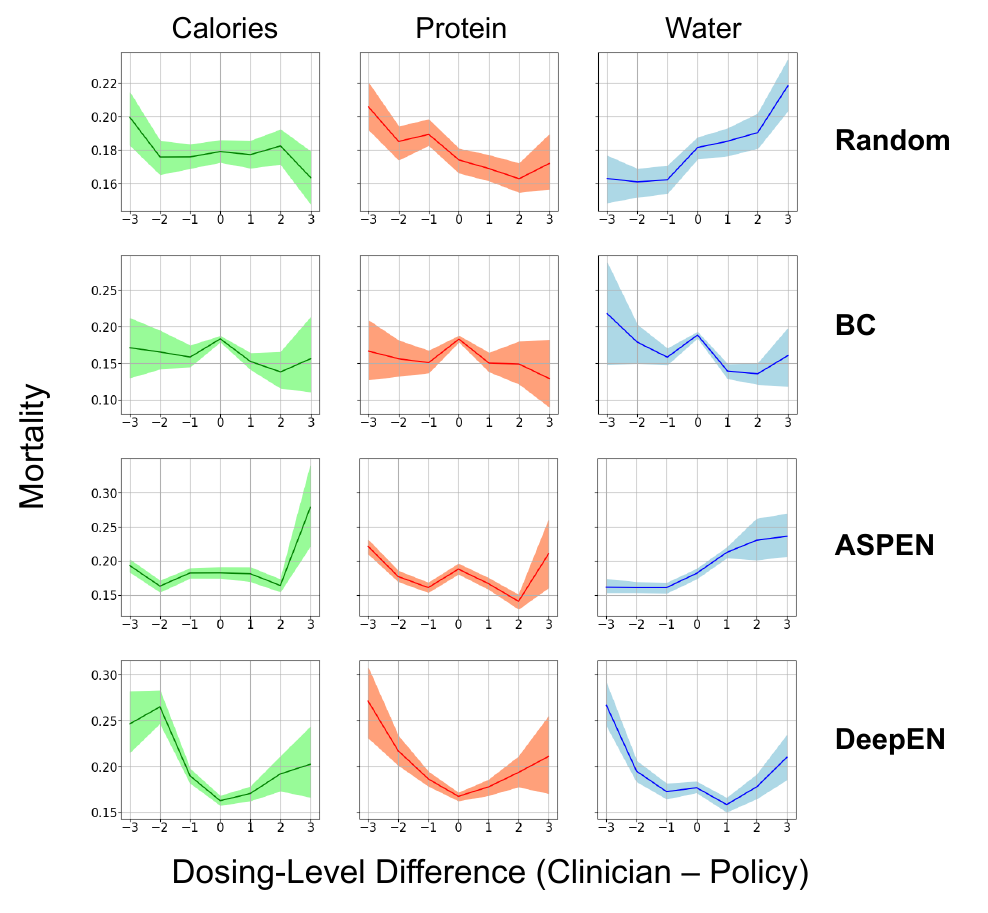} 
\caption{Dosing differences (x-axis) versus mortality (y-axis) for all policies. 
The shaded area indicates the confidence interval.}
\label{fig:vplot_mort}
\vspace{-0.5em} 
\end{figure*}

We next examined how divergence between policy recommendations and clinician actions relates to outcomes. Figure~\ref{fig:vplot_mort} displays the relationship between action-level differences and mortality across calorie, protein, and water components. For DeepEN, mortality is lowest when model and clinician recommendations are closely aligned and increases with larger discrepancies. This structured pattern suggests that the learned policy tends to agree with clinician practice in lower-risk regions while diverging more substantially in regions associated with higher mortality.

\begin{figure*}[!t]
\centering
\includegraphics[width=1\textwidth]{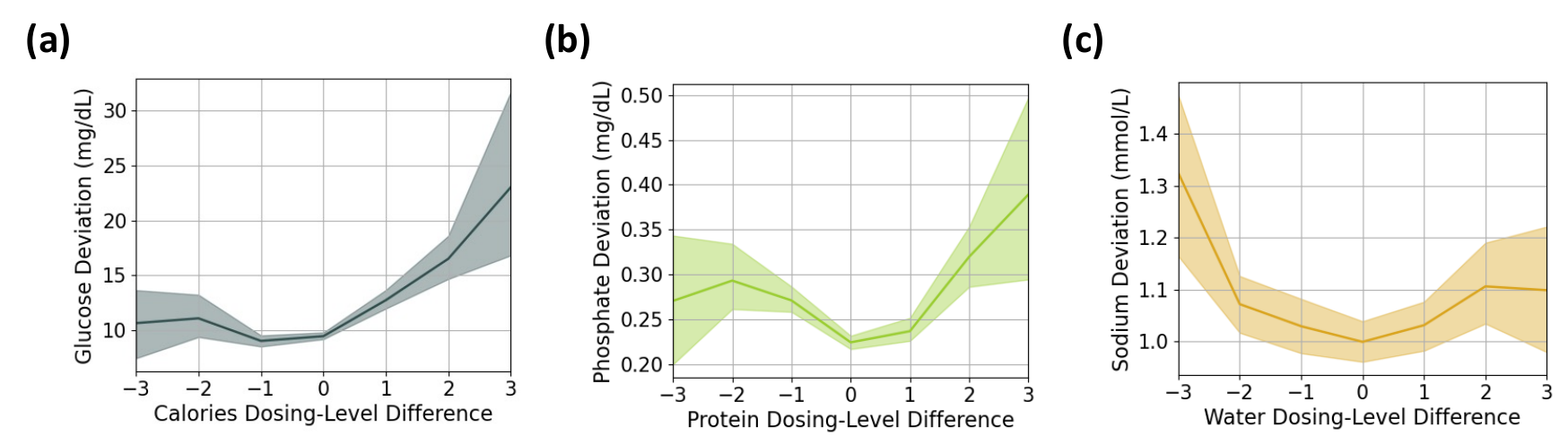} 
\caption{Dosage differences versus biomarker deviation for DeepEN. 
(a) Glucose vs.\ calories difference, 
(b) Phosphate vs.\ protein difference, and 
(c) Sodium vs.\ water difference.}
\label{fig:vplot_biomarker}
\vspace{-0.5em} 
\end{figure*}

A similar structural relationship is observed for metabolic stability (Figure~\ref{fig:vplot_biomarker}). For glucose, phosphate, and sodium, biomarker deviation is minimized near action concordance and increases with larger discrepancies. These trends indicate that policy disagreement is systematically associated with physiologic instability rather than occurring randomly.

Because patients with low discrepancy may simply represent lower baseline severity, we performed multivariable regression analyses adjusting for age, SOFA score, and weight (Supplementary Table~S4). After adjustment, greater deviation from the DeepEN policy remained independently associated with increased mortality and biomarker instability, though causal inference cannot be established in this retrospective setting. In contrast, deviation from a random policy showed no such association; this suggests that the learned policy captures outcome-relevant treatment structure rather than arbitrary action patterns.

\subsection{Model Interpretability}

The feature importance analysis indicates that both clinician and DeepEN policies are anchored in clinically fundamental determinants of nutrition dosing, with patient body weight and time since enteral nutrition (EN) initiation emerging as the dominant drivers across calorie, protein, and water decisions (Figure~\ref{fig:feature_importance}). This is clinically expected: body weight underlies baseline caloric and protein requirements in established guidelines, while time since EN initiation reflects staged feeding progression, tolerance assessment, and gradual advancement practices in critical care \cite{adeyinka2018enteric,preiser2021guide}.

Beyond these shared determinants, the DeepEN policy demonstrates a broader reliance on markers of organ function and physiologic stability, particularly urine output, creatinine, shock index, and electrolyte measures such as phosphate and sodium. These variables are consistently represented across dosing domains, suggesting that the learned policy incorporates signals of renal function, hemodynamic stability, and metabolic balance when determining nutrient and fluid delivery. This pattern is consistent with emerging critical care nutrition evidence emphasizing the importance of tailoring protein and fluid provision according to dynamic organ function, renal replacement status, and metabolic tolerance rather than relying solely on static weight-based targets \cite{wischmeyer2023personalized,dresen2023clinician}. In contrast, clinician policies appear more concentrated on weight- and time-based cues, with comparatively lower emphasis on dynamic organ-function indicators.

The prominence of urine output across all three nutrition components is particularly notable, as it serves as an integrated marker of renal function, volume status, and overall physiologic stability—factors that directly influence fluid administration and indirectly shape protein and caloric tolerance. Recent literature highlights the need to individualize nutrition strategies in the context of evolving kidney function, fluid balance, and risk of electrolyte disturbances, particularly during the early phase of critical illness and refeeding \cite{mehanna2008refeeding,da2020aspen}. Collectively, these findings suggest that DeepEN moves beyond static weight-based heuristics and instead operationalizes a physiology-informed approach aligned with contemporary perspectives on personalized nutrition in the ICU \cite{kittrell2024role}, consistent with its longer-horizon optimization objective.

\begin{figure*}[!t]
\centering
\includegraphics[width=1\textwidth]{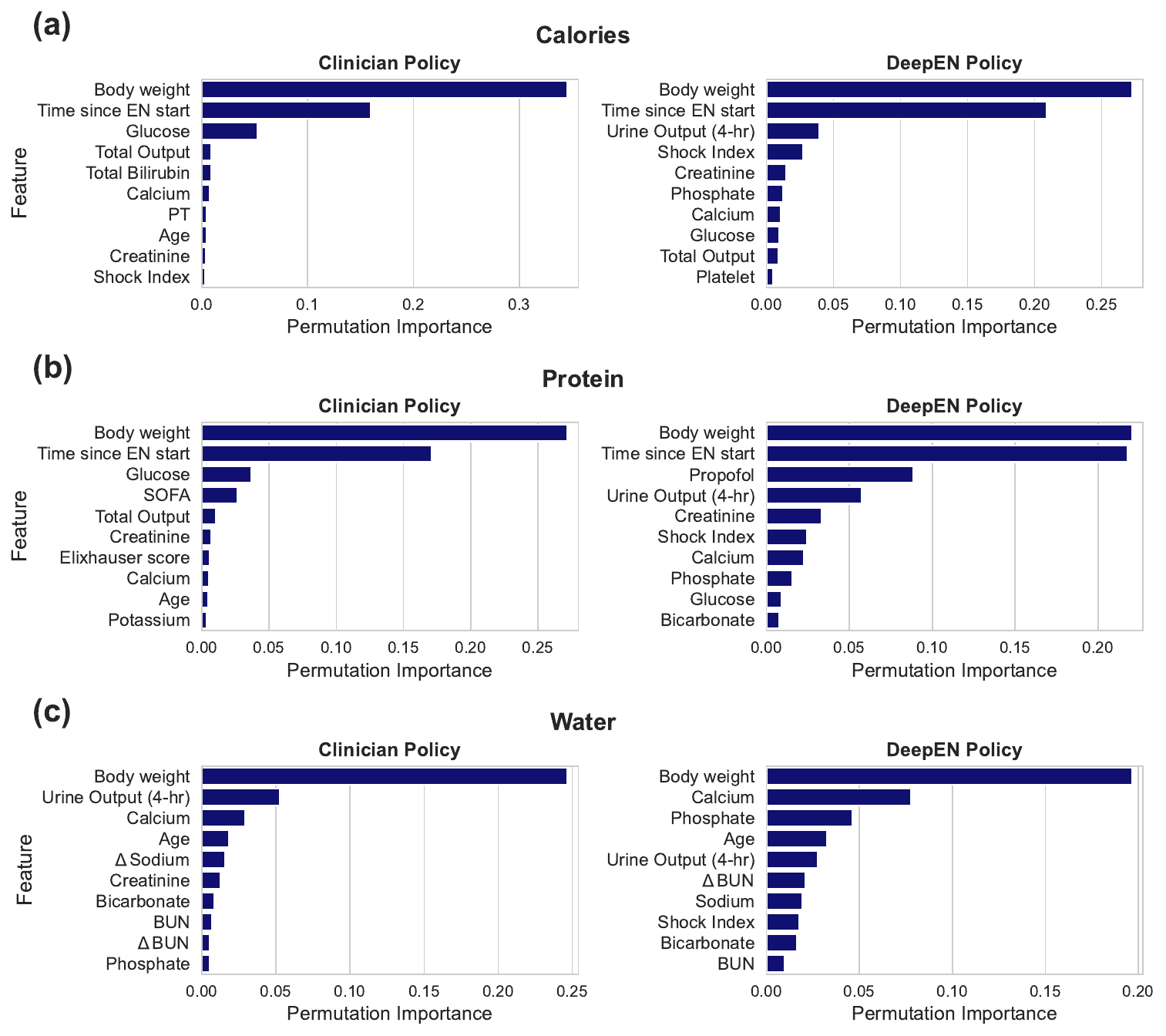} 
\caption{Top 10 most influential features are shown for calorie (a), protein (b), and water (c) dosing decisions under the clinician policy (left) and the DeepEN policy (right) based on permutation importance.}
\label{fig:feature_importance}
\vspace{-0.5em} 
\end{figure*}

Similarly, the joint action heatmaps (Supplementary Figure~S2) indicate that DeepEN learns more structured and coordinated enteral nutrition strategies than observed clinician practice. Clinician policies exhibit conservative low–low dosing patterns early in the feeding course followed by largely parallel escalation of calories, protein, and water. In contrast, DeepEN demonstrates a more graded progression, including partial decoupling of protein from calorie escalation and tighter alignment between nutritional intensity and water provision. These patterns suggest that the learned policy captures coordinated treatment structure beyond static rule-based advancement.

\FloatBarrier

\section{Discussion and Conclusion}

DeepEN introduces a reinforcement learning framework for multi-component enteral nutrition optimization in critically ill patients, targeting improved hospital survival while preserving metabolic stability. This work establishes a principled strategy for applying conservative offline RL to a previously unmodeled clinical decision structure characterized by interdependent dosing components, delayed physiologic effects, and sparse long-term outcomes \cite{raghu2017deep,lu2024reinforcement}. By integrating physiologically grounded reward shaping, conservative regularization \cite{kumar2020conservative}, and structured off-policy evaluation, DeepEN provides a reproducible foundation for safe policy learning in this problem class.

Across multiple evaluation paradigms, DeepEN demonstrated consistent improvements in estimated survival and metabolic stability, supporting the robustness of its learned policy structure. Its improvement relative to behavior cloning suggests that explicit long-horizon optimization provides benefits beyond simple imitation of historical clinician decisions \cite{raghu2017continuous,wu2023value}. These improvements were corroborated through return-calibrated mortality and biomarker stability analyses, reinforcing that performance gains reflect clinically meaningful outcome alignment rather than artifacts of reward design.

Interpretability analyses further support the validity of the learned policy. Rather than focusing solely on variable-level importance, we examined coordinated treatment structure within the multi-dimensional action space \cite{tan2024advancing}. The results indicate that DeepEN operates through integrated treatment patterns that adapt to dynamic physiologic context, suggesting that the model captures higher-order structure in the data rather than isolated correlations.

Beyond the EN application itself, this study provides a methodological foundation for multi-component, slow-acting clinical treatment optimization under offline constraints. Nutritional management differs fundamentally from rapid-response hemodynamic control in that its effects are metabolically mediated and temporally diffuse. The adaptations introduced here—including physiologically aligned intermediate rewards, return–outcome calibration, and policy-level analysis in combinatorial action spaces—offer a transferable framework for other biomedical domains involving coordinated therapies and delayed feedback. This design also avoids evaluation restricted to clinician-policy agreement regions, which can induce selection-on-agreement bias and overstate policy performance \cite{gottesman2018evaluating}.

From an implementation standpoint, DeepEN is designed as a clinician-facing decision support tool within a human-in-the-loop paradigm \cite{topol2019high,sendak2020real}. Because the model relies exclusively on routinely collected structured ICU data, integration into electronic health record systems is technically feasible through standard interoperability frameworks. Recommendations would be presented alongside contextual trends, with final decisions remaining under clinical supervision. Prospective validation through silent-mode deployment or simulation-based studies will be necessary to assess real-world usability and clinician–AI interaction.

Several limitations warrant consideration. Although the cohort size is substantial, data were derived from a single hospital system, potentially limiting cross-institutional generalizability. External validation will therefore be essential. At present, no other open-source critical care datasets provide sufficiently granular and longitudinal enteral nutrition delivery data—including detailed calorie, protein, and fluid dosing trajectories—to enable comparable replication. This reflects a broader structural limitation in publicly available ICU datasets \cite{johnson2023mimic}. In addition, the reward formulation, while physiologically aligned and deliberately simplified to reduce parameter sensitivity, remains handcrafted \cite{raghu2017deep,lin2018deep,eghbali2021patient} and may not capture all dimensions of nutritional adequacy.

In summary, DeepEN demonstrates that conservative offline reinforcement learning can support personalized optimization of multi-component enteral nutrition in the ICU. By combining long-horizon optimization with explicit safety constraints and interpretable evaluation, the framework illustrates the potential for data-driven decision support systems to augment guideline-based care in complex clinical therapies. Continued multi-institutional validation and prospective evaluation will be essential for safe clinical translation.

\section{Ethical Statement}
In line with prior work on reinforcement learning–driven clinical decision support systems (CDSSs) for conditions such as sepsis \cite{komorowski2018artificial}, our proposed framework adopts a human-in-the-loop approach. The system is designed to assist, rather than replace, the role of clinicians by offering data-driven recommendations derived from past clinical trajectories. Importantly, the ultimate responsibility for treatment decisions remains with the clinician, preserving professional autonomy and ensuring that medical expertise governs final care plans. Our approach enables the integration of machine intelligence with clinical judgment, leveraging the strengths of both to improve patient outcomes without undermining ethical standards.

Additionally, our models were developed and validated solely on de-identified, publicly available retrospective data, with no involvement of real-time clinical interventions or patient contact. As such, the work poses no direct risk to patients and aligns with ethical research practices, minimizing concerns regarding safety, accountability, or consent in the current study phase.

\section{Data Availability}
The open-source MIMIC-IV 2.0 data used in this study is available at 
\url{https://physionet.org/content/mimiciv/2.0/}.

\section{Code Availability}
The underlying code for this study is available on Github via \url{https://github.com/danjst/DeepEN}.

\section{Acknowledgment}
The authors would like to acknowledge the  contributions of their team members and mentors from the Singapore IMAGINE AI 2024 Datathon.

\section{Declaration of generative AI and AI-assisted technologies in the manuscript preparation process
}
During the preparation of this work, the authors used ChatGPT (OpenAI, San Francisco, CA, USA) to assist with language refinement, grammatical editing, and improvement of clarity in selected sections of the manuscript. After using this tool, the authors reviewed and edited the content as needed and take full responsibility for the content of the published article.

\bibliographystyle{elsarticle-num}
\bibliography{references}

\section{Funding}
This research is supported by A*STAR, CISCO Systems (USA) Pte. Ltd., and the National University of Singapore under its Cisco-NUS Accelerated Digital Economy Corporate Laboratory (Award I21001E0002),  the National University of Singapore President's Graduate Fellowship, and the AI for Public Health Program in Saw Swee Hock School of Public Health. 

\section{Author Contributions}
MF is the guarantor of the content of the manuscript, including the data and analysis. KCS conceptualized and supervised the study, and aided in data interpretation and manuscript editing. DJT designed the algorithm, collected data and results, and was the primary manuscript writer. JC contributed to data analysis, interpretation, and manuscript writing. DP assisted in editing of the manuscript. All authors participated in manuscript review, provided final approval of the manuscript, and take responsibility for the accuracy and integrity of the work. 


\clearpage

\setcounter{section}{0}
\setcounter{subsection}{0}
\setcounter{subsubsection}{0}

\renewcommand{\thesection}{\arabic{section}}
\renewcommand{\thesubsection}{\thesection.\arabic{subsection}}
\renewcommand{\thesubsubsection}{\thesubsection.\arabic{subsubsection}}

\vspace{0.5em}
\section*{\fontsize{16}{18}\selectfont\bfseries Supplementary Information}
\vspace{0.5em}

\addcontentsline{toc}{section}{Supplementary Information}

\renewcommand{\thefigure}{S\arabic{figure}}
\renewcommand{\thetable}{S\arabic{table}}
\setcounter{figure}{0}
\setcounter{table}{0}
\setlength{\tabcolsep}{4pt}     
\renewcommand{\arraystretch}{0.95}

\setlength{\tabcolsep}{4pt}
\renewcommand{\arraystretch}{0.95}

\captionsetup{justification=centering}
\captionsetup[table]{skip=4pt}

\makeatletter
\renewcommand{\section}{\@startsection{section}{1}{\z@}%
  {-2.5ex \@plus -1ex \@minus -.2ex}%
  {1.5ex \@plus .2ex}%
  {\normalfont\large\bfseries}}

\renewcommand{\subsection}{\@startsection{subsection}{2}{\z@}%
  {-2.0ex \@plus -1ex \@minus -.2ex}%
  {1.0ex \@plus .2ex}%
  {\normalfont\normalsize\itshape}}
\makeatother

\section*{Supplementary Methods}

\subsubsection*{ASPEN-Derived Policy Definition}

These nutritional targets are based on the 2021 ASPEN requirements for enteral nutrition \cite{compher2022guidelines}. Protein targets were modestly adapted based on more recent evidence recommending a low-dosing period during the early-acute phase of critical illness \cite{stoian2025personalized,wischmeyer2023personalized}. 

\subsubsection*{Calories}
\begin{itemize}
    \item For patients with BMI $< 30$: provide \textbf{25 kcal/kg/day}.
    \item For patients with BMI between $30$ and $50$: provide \textbf{22 kcal/kg/day}.
    \item For patients with BMI $> 50$: provide \textbf{11 kcal/kg/day}.
    \item For all patients, only \textbf{70\%} of the above amounts are provided during the \textbf{first 3 days} of enteral feeding to simulate hypocaloric underfeeding.
\end{itemize}

\subsubsection*{Protein} 
\textit{Early Acute Phase (ICU Days 1--4)}
\begin{itemize}
    \item Days 1--2: \textbf{0.8 g/kg/day} of protein.
    \item Days 3--4: \textbf{1.0 g/kg/day} of protein.
\end{itemize}

\textit{Stable Phase (ICU Days $>$4)}
\begin{itemize}
    \item For patients with BMI $< 30$: provide \textbf{1.2 g/kg/day} of protein.
    \item For patients with BMI between $30$ and $40$: provide \textbf{2 g/kg ideal body weight/day}.
    \item For patients with BMI $> 40$: provide \textbf{2.5 g/kg ideal body weight/day}.
    \item Patients with \textbf{burns} receive \textbf{2 g/kg/day} of protein, regardless of BMI.
    \item Patients on \textbf{CRRT} receive \textbf{2.5 g/kg/day} of protein, regardless of BMI.
\end{itemize}

\subsubsection*{Water}
\begin{itemize}
    \item 1.5 ml per 1 kcal of calories administered.
\end{itemize}

\section*{Supplementary Tables}

\begin{table}[ht]
\centering

\caption{Action Quantile Thresholds (4-hourly doses)}
\begin{tabular}{lcccc}
\toprule
\textbf{Quantile} & \textbf{1} & \textbf{2} & \textbf{3} & \textbf{4} \\
\midrule
\textbf{Calories (kcal/kg)} & 0--1.91 & 1.91--3.05 & 3.05--4.13 & $>$4.13 \\
\textbf{Protein (g/kg)}     & 0--0.08 & 0.08--0.14 & 0.14--0.19 & $>$0.19 \\
\textbf{Water (ml/kg)}      & 0--3.61 & 3.61--5.40 & 5.40--8.15 & $>$8.15 \\
\bottomrule
\end{tabular}

\label{tab:quantile_thresholds}
\end{table}

\begin{table}[ht]
\centering
\caption{Cohort Characteristics}
\label{tab:cohort_characteristics}
\begin{tabular}{lcccc}
\toprule
\textbf{Cohort} & \textbf{\% Female} & \textbf{Age, years} & \textbf{ICU Stay, hours} & \textbf{Total (n)} \\
 &  & \textbf{Median (IQR)} & \textbf{Median (IQR)} &  \\
\midrule
Overall 
& 41.63 
& 67 (55--77) 
& 181.7 (99.2--311.9) 
& 11378 \\

Non-Survivors 
& 43.19 
& 70 (59--80) 
& 198.1 (120.1--319.9) 
& 2556 \\

Survivors 
& 41.18 
& 65 (54--76) 
& 175.5 (93.8--309.1) 
& 8822 \\

\bottomrule
\end{tabular}
\end{table}

\small
\setlength{\tabcolsep}{6pt}

\begin{longtable}{ccccr}
\caption{Distribution of Discrete Actions in the Dataset}
\label{tab:action_distribution} \\

\toprule
\textbf{ID} & \textbf{Cal} & \textbf{Pro} & \textbf{Water} & \textbf{Count (\%)} \\
\midrule
\endfirsthead

\caption[]{(continued)} \\
\toprule
\textbf{ID} & \textbf{Cal} & \textbf{Pro} & \textbf{Water} & \textbf{Count (\%)} \\
\midrule
\endhead

\bottomrule
\endfoot

0 & 1 & 1 & 1 & 21268 (9.5) \\
1 & 4 & 4 & 4 & 15932 (7.1) \\
2 & 4 & 4 & 3 & 13241 (5.9) \\
3 & 2 & 2 & 1 & 11755 (5.2) \\
4 & 4 & 4 & 2 & 9322 (4.2) \\
5 & 3 & 3 & 2 & 9179 (4.1) \\
6 & 1 & 1 & 2 & 9000 (4.0) \\
7 & 2 & 2 & 2 & 8683 (3.9) \\
8 & 1 & 1 & 4 & 8392 (3.7) \\
9 & 1 & 1 & 3 & 8093 (3.6) \\
10 & 3 & 3 & 3 & 7969 (3.6) \\
11 & 2 & 2 & 3 & 6763 (3.0) \\
12 & 3 & 3 & 1 & 5740 (2.6) \\
13 & 2 & 2 & 4 & 5536 (2.5) \\
14 & 3 & 3 & 4 & 5458 (2.4) \\
15 & 4 & 3 & 4 & 5169 (2.3) \\
16 & 3 & 4 & 2 & 4777 (2.1) \\
17 & 2 & 3 & 2 & 4695 (2.1) \\
18 & 3 & 4 & 3 & 4644 (2.1) \\
19 & 1 & 2 & 1 & 4325 (1.9) \\
20 & 4 & 3 & 3 & 4253 (1.9) \\
21 & 3 & 2 & 4 & 4098 (1.8) \\
22 & 2 & 3 & 1 & 3993 (1.8) \\
23 & 3 & 2 & 3 & 3485 (1.6) \\
24 & 2 & 1 & 4 & 3434 (1.5) \\
25 & 3 & 4 & 4 & 3266 (1.5) \\
26 & 3 & 2 & 2 & 3147 (1.4) \\
27 & 2 & 3 & 3 & 3128 (1.4) \\
28 & 4 & 3 & 2 & 2980 (1.3) \\
29 & 3 & 2 & 1 & 2352 (1.1) \\
30 & 4 & 4 & 1 & 2337 (1.0) \\
31 & 2 & 1 & 3 & 2232 (1.0) \\
32 & 1 & 2 & 2 & 1955 (0.9) \\
33 & 2 & 3 & 4 & 1774 (0.8) \\
34 & 2 & 1 & 2 & 1578 (0.7) \\
35 & 1 & 2 & 3 & 1384 (0.6) \\
36 & 4 & 3 & 1 & 1218 (0.5) \\
37 & 2 & 1 & 1 & 1205 (0.5) \\
38 & 3 & 4 & 1 & 1179 (0.5) \\
39 & 1 & 2 & 4 & 1062 (0.5) \\
40 & 4 & 2 & 4 & 1037 (0.5) \\
41 & 3 & 1 & 4 & 535 (0.2) \\
42 & 2 & 4 & 1 & 385 (0.2) \\
43 & 2 & 4 & 2 & 368 (0.2) \\
44 & 4 & 2 & 3 & 324 (0.1) \\
45 & 2 & 4 & 3 & 311 (0.1) \\
46 & 1 & 3 & 1 & 272 (0.1) \\
47 & 2 & 4 & 4 & 249 (0.1) \\
48 & 3 & 1 & 3 & 163 (0.1) \\
49 & 4 & 2 & 2 & 162 (0.1) \\
50 & 1 & 3 & 2 & 147 (0.1) \\

\end{longtable}

\begin{table}[ht]
\centering
\normalsize
\setlength{\tabcolsep}{3pt}
\renewcommand{\arraystretch}{1.05}

\caption{Multivariable Regression Analyses of Distance from Clinician's Policy and Clinical Outcomes. Coefficients shown as $\beta$ (95\% CI).}
\label{tab:regression_appendix}
\makebox[\textwidth][c]{%
\begin{tabular}{lcccc}
\toprule
& DeepEN & ASPEN & BC & Random \\
\midrule
\multicolumn{5}{l}{\textbf{A. Mortality}} \\
Distance 
& 1.06*** (0.67, 1.45) 
& 0.30* (0.00, 0.60) 
& -0.86** (-1.44, -0.27) 
& -0.06 (-0.51, 0.40) \\
Age 
& 0.03*** (0.02, 0.04) 
& 0.03*** (0.02, 0.03) 
& 0.03*** (0.02, 0.03) 
& 0.03*** (0.02, 0.03) \\
Gender 
& 0.19 (-0.03, 0.41) 
& 0.18 (-0.04, 0.39) 
& 0.22* (0.01, 0.44) 
& 0.19 (-0.02, 0.41) \\
SOFA 
& 0.18*** (0.15, 0.21) 
& 0.17*** (0.14, 0.20) 
& 0.18*** (0.15, 0.20) 
& 0.18*** (0.15, 0.21) \\
Weight 
& 0.00 (-0.01, 0.00) 
& -0.01** (-0.01, 0.00) 
& -0.01*** (-0.01, 0.00) 
& -0.01** (-0.01, 0.00) \\
\midrule
\multicolumn{5}{l}{\textbf{B. Glucose Deviation}} \\
Distance 
& 9.96*** (6.16, 13.76) 
& 0.22 (-2.71, 3.14) 
& 1.58 (-3.70, 6.85) 
& 1.94 (-2.60, 6.48) \\
Age 
& 0.09** (0.02, 0.15) 
& 0.07* (0.01, 0.14) 
& 0.07* (0.01, 0.14) 
& 0.07* (0.01, 0.14) \\
Gender 
& -1.84 (-3.87, 0.19) 
& -1.75 (-3.80, 0.30) 
& -1.80 (-3.85, 0.25) 
& -1.74 (-3.78, 0.31) \\
SOFA 
& 0.28* (0.01, 0.55) 
& 0.30* (0.02, 0.58) 
& 0.31* (0.04, 0.59) 
& 0.30* (0.03, 0.58) \\
Weight 
& 0.10*** (0.05, 0.14) 
& 0.08*** (0.03, 0.12) 
& 0.08*** (0.04, 0.12) 
& 0.08*** (0.03, 0.12) \\
\midrule
\multicolumn{5}{l}{\textbf{C. Phosphate Deviation}} \\
Distance 
& 0.27*** (0.17, 0.37) 
& 0.03 (-0.04, 0.11) 
& -0.15* (-0.29, -0.02) 
& -0.09 (-0.21, 0.04) \\
Age 
& 0.00 (0.00, 0.00) 
& 0.00 (0.00, 0.00) 
& 0.00 (0.00, 0.00) 
& 0.00 (0.00, 0.00) \\
Gender 
& 0.00 (-0.05, 0.05) 
& 0.00 (-0.05, 0.05) 
& 0.01 (-0.04, 0.06) 
& 0.01 (-0.05, 0.06) \\
SOFA 
& 0.03*** (0.03, 0.04) 
& 0.03*** (0.03, 0.04) 
& 0.03*** (0.03, 0.04) 
& 0.03*** (0.03, 0.04) \\
Weight 
& 0.00* (0.00, 0.00) 
& 0.00 (0.00, 0.00) 
& 0.00 (0.00, 0.00) 
& 0.00 (0.00, 0.00) \\
\midrule
\multicolumn{5}{l}{\textbf{D. Sodium Deviation}} \\
Distance 
& 1.20*** (0.81, 1.59) 
& 0.06 (-0.23, 0.34) 
& -0.13 (-0.65, 0.39) 
& 0.12 (-0.38, 0.62) \\
Age 
& 0.03*** (0.02, 0.04) 
& 0.03*** (0.02, 0.03) 
& 0.03*** (0.02, 0.03) 
& 0.03*** (0.02, 0.03) \\
Gender 
& 0.19 (-0.03, 0.41) 
& 0.18 (-0.04, 0.39) 
& 0.22* (0.01, 0.44) 
& 0.19 (-0.02, 0.41) \\
SOFA 
& 0.18*** (0.15, 0.21) 
& 0.17*** (0.14, 0.20) 
& 0.18*** (0.15, 0.20) 
& 0.18*** (0.15, 0.21) \\
Weight 
& 0.00 (-0.01, 0.00) 
& -0.01** (-0.01, 0.00) 
& -0.01*** (-0.01, 0.00) 
& -0.01** (-0.01, 0.00) \\
\bottomrule
\end{tabular}
}

\begin{center}
\footnotesize
\textit{Notes:} * $p < 0.05$, ** $p < 0.01$, *** $p < 0.001$. All models adjusted for age, gender, SOFA score, and weight.
\end{center}
\end{table}

\FloatBarrier

\begin{table}[ht]
\centering
\normalsize
\setlength{\tabcolsep}{6pt}
\caption{Sensitivity analysis of intermediate reward scaling coefficient $\lambda$}
\label{tab:lambda_sensitivity}
\begin{tabular}{cc}
\toprule
$\lambda$ & Mortality (\%, 95\% CI) \\
\midrule
0.0 & 22.1 $\pm$ 1.3 \\
0.1 & 19.2 $\pm$ 0.8 \\
\textbf{0.2} & \textbf{18.8 $\pm$ 1.0} \\
0.3 & 19.5 $\pm$ 1.0 \\
0.4 & 21.2 $\pm$ 1.1 \\
0.5 & 22.0 $\pm$ 1.2\\
\bottomrule
\end{tabular}

\begin{flushleft}
\centering
\footnotesize
\textit{Note:} Mortality reported as mean $\pm$ 95\% CI half-width.
\end{flushleft}

\end{table}


\section*{Supplementary Figures}

\begin{figure}[!htbp]
\centering
\includegraphics[width=1\textwidth]{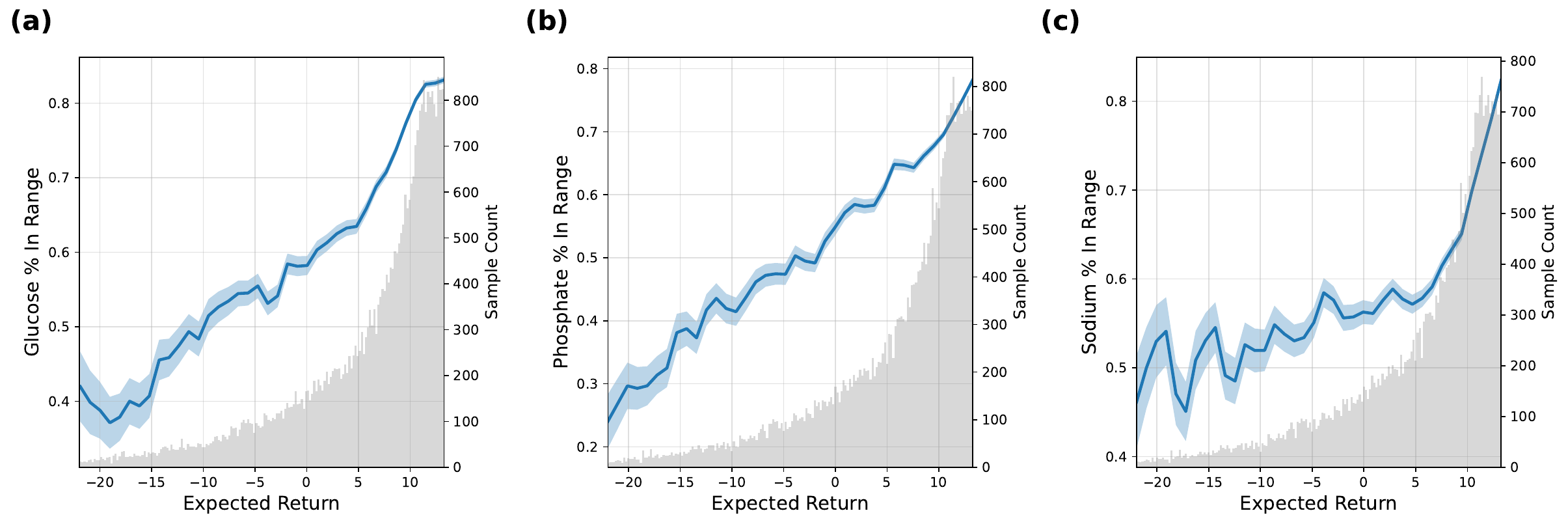} 
\caption{Relationship between expected return and biomarker stability for (a) glucose, (b) phosphate, and (c) sodium on the held-out test set. Shaded regions indicate 95\% confidence intervals, and gray histograms show the sample distribution across expected return values.}
\label{fig:qplot_biomarker}
\vspace{-0.5em} 
\end{figure}

\begin{figure}[!htbp]
\centering
\includegraphics[width=1\textwidth]{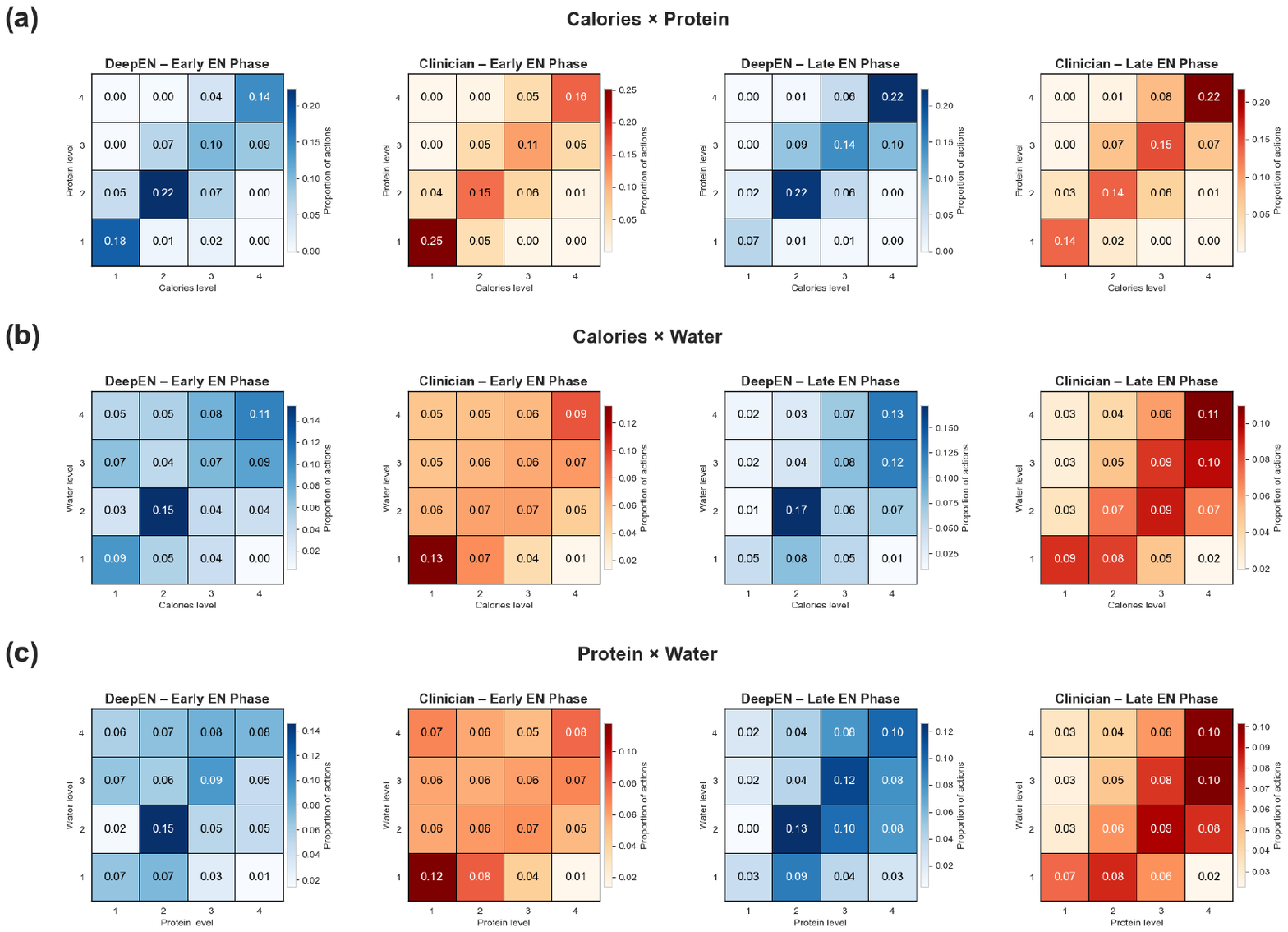} 
\caption{Pairwise heatmaps of discrete enteral nutrition actions selected by the DeepEN policy and clinicians for (a) calories $\times$ protein, (b) calories $\times$ water, and (c) protein $\times$ water, stratified into early ($\leq$3 days) and late ($>$3 days) feeding phases.}
\label{fig:heatmap}
\end{figure}



\end{document}